# Lensless Photography with only an image sensor

Ganghun Kim,[1] Kyle Isaacson,[2] Racheal Palmer,[2] Rajesh Menon[1,a)]

[1]*Department of Electrical and Computer Engineering, University of Utah, Salt Lake City, 84111, USA*

[2]*Department of Bioengineering, University of Utah, Salt Lake City, 84111, USA*

*\*Corresponding author: rmenon@eng.utah.edu*

Photography usually requires optics in conjunction with a recording device (an image sensor). Eliminating the optics could lead to new form factors for cameras. Here, we report a simple demonstration of imaging using a bare CMOS sensor that utilizes computation. The technique relies on the space variant point-spread functions resulting from the interaction of a point source in the field of view with the image sensor. These space-variant point-spread functions are combined with a reconstruction algorithm in order to image simple objects displayed on a discrete LED array as well as on an LCD screen. We extended the approach to video imaging at the native frame rate of the sensor. Finally, we performed experiments to analyze the parametric impact of the object distance. Improving the sensor designs and reconstruction algorithms can lead to useful cameras without optics.

The optical systems of cameras in mobile devices typically constrain the overall thickness of the devices[1-2]. By eliminating the optics, it is possible to create ultra-thin cameras with interesting new form factors. Previous work in computational photography has eliminated the need for lenses by utilizing apertures in front of the image sensor[3-7] or via coherent illumination of the sample[8]. In the former case, the apertures create shadow patterns on the sensor that could be computationally recovered by solving a linear inverse problem[9]. The latter case requires coherent illumination, which is not generally applicable to imaging. In most instances, coded apertures have replaced the lenses. Microfabricated coded apertures have recently shown potential for the thinner systems,[4] with thickness on the order of millimeters. However, these apertures are absorbing and hence, exhibit relatively low transmission efficiencies. Another method utilizes holographic phase masks integrated onto the image sensor in conjunction with computation to enable simple imaging.[10,11] In this case, precise microfabrication of the mask onto the sensor is required. Another computational camera utilizes a microlens array to form a large number of partial images of the scene, which is then numerically combined to form a single image with computational refocusing[12-14]. Here, we report on a computational camera that is comprised of only a conventional image sensor and no other elements.

Our motivation for this camera is based upon the recognition that all cameras essentially rely on the fact that the information about the object enters the aperture of the lens, the coded aperture or micro-lens array, and is recorded by the image sensor. In the case of the coded aperture and the microlens array, numerical processing is performed to represent the image for human consumption. If all optical elements are eliminated, the information from the object is still recorded by the image sensor. If appropriate reconstruction algorithms were developed, the image recorded by the sensor can be subsequently recovered for human consumption. It is analogous the multi-sensory compressive sensing[6] where spatial light

---

[a)] Correspondence and requests for materials should be addressed to R. M. (rmenon@eng.utah.edu)

modulator is used to form multiple arbitrary apertures in front of sensor array to record the scene, except the aperture in this case is a single static open aperture.

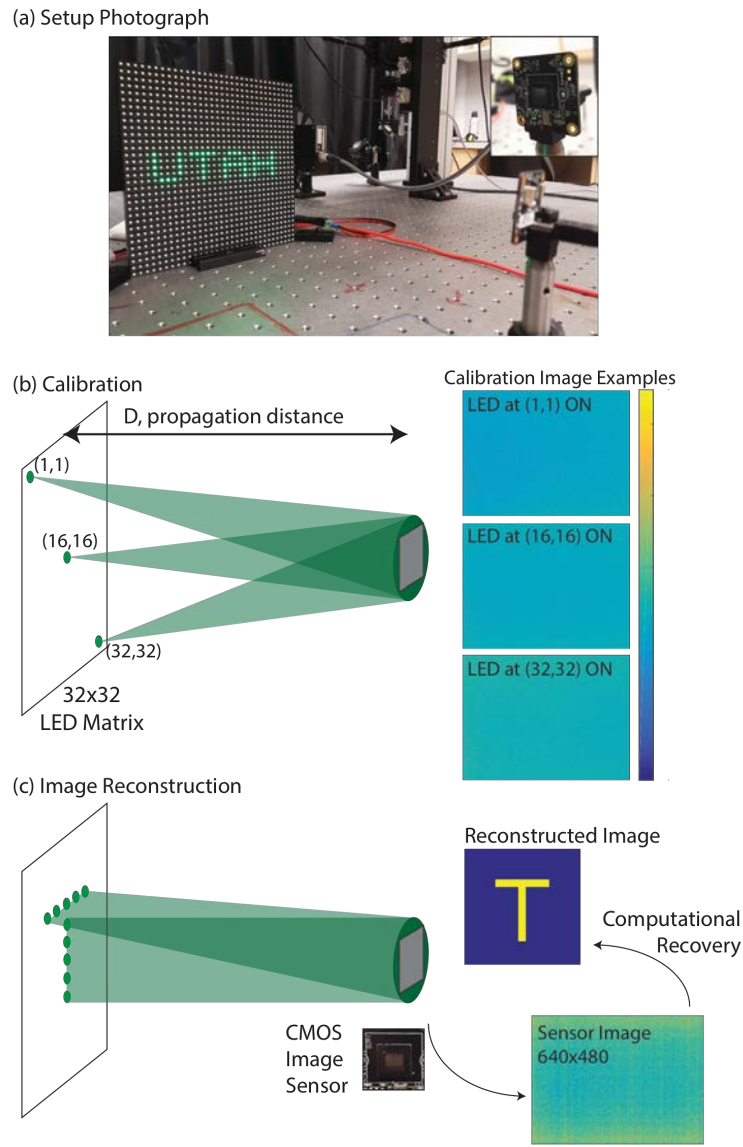

FIG. 1. Experiment setup and principles of operation. (a) Photograph of the setup showing LED matrix in front of the image sensor (top-right inset). (b) Sensor images are recorded for each LED in the array separately. Three exemplary calibration images are shown at three locations of the LED point source. (c) To test image reconstruction, a simple pattern is displayed on the LED matrix. The image captured by the sensor is input into the recovery algorithm to reconstruct the scene.

Our test object was a 32x32 RGB LED matrix panel (SparkFun, COM-12584) controlled via an Arduino board. The LEDs are uniformly spaced with a gap of 6.1mm. The total matrix size is 19cm X 19cm. The image sensor was a commercially available CMOS color sensor (DFM

22BUC03-LM, Imaging Source). The sensor has 640 x 480 pixels, each of which is a 6μm square. The maximum supported frame rate of the camera via a USB 2.0 interface is 76 frames per second. A photograph of our setup is shown in Fig. 1(a). The sensor was placed at a set distance, D from the object as illustrated in Fig. 1(b). The refresh rate of the LED array is 16Hz. We used frame averaging to eliminate the effect of this refresh rate as well as to enhance the signal to noise ratio of the raw data.

As the first step, we experimentally calibrated the imaging system by recording the image of each LED in the array, forming what we refer to as the calibration matrix, *A*. 100 frames were averaged for each LED. Exemplary frames (elements of matrix, *A*) are shown in Fig. 1(b). Extraneous light was minimized during all experiments. When an arbitrary pattern is displayed on the LED matrix, the resulting sensor image is a linear combination of the images formed by the individual LEDs (elements of the calibration matrix, *A*). The system can be described as

$$b = Ax \qquad (1)$$

where *x* is the object, and *b* is the sensor image. Calibration matrix A collected in this setup has condition number of 10,912, which suggests the posed inverse problem is ill-posed, and therefore mostly likely fail if solved using a direct matrix inversion. Hence, we can attempt to solve a regularized linear inverse problem to recover the object *x* from the sensor data *b*, as formulated below:

$$\hat{x} = \mathrm{argmin} \|Ax - b\|_2^2 + \alpha^2 \|x\|_2^2 \qquad (2)$$

where argmin refers to the L2 minimization. The regularization parameter, **α** controls noise suppression and the smoothness of the solution, *x*. Regularization provides robust reconstruction capability that can withstand reasonable level of noise, as evidenced by the following demonstrations of the experimental condition.

Since our LED matrix has green LEDs, we used only the green channel in our experiments. But our technique is easily extended to all colors. Note that the exposure time was chosen between 10ms and 100ms to ensure that the sensor pixels were not saturated. For each image, 100 frames were averaged. An example of the raw sensor image and the reconstructed image are shown in Fig. 1(c), where the object is the letter, "T". More interesting objects are illustrated in Fig. 2, where the first column shows the photographs of the LED matrix using a conventional camera, the second column shows the raw sensor images, and the third and fourth columns show the reconstructed images before and after thresholding.

Using this same technique, we can perform video imaging as well, as illustrated by an example video of a jumping stick-man included as supplementary video 1. The reconstruction process, including all necessary image processing, takes less than 10ms per frame using a regular desktop PC (Intel Core i7-4790, 32 GB memory). In our current implementation, the frame rate was limited by our averaging of 100 frames, which takes 1 to 10 seconds.

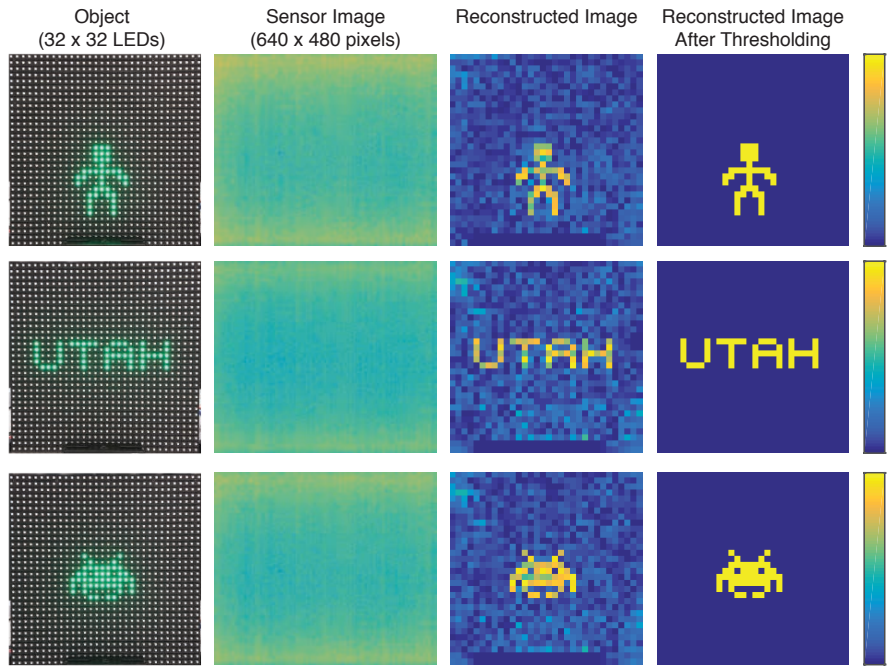

FIG. 2. Exemplary images taken with our sensor. The left-column shows the objects displayed on the LED matrix. The 2nd column shows the raw sensor images. The 3rd column shows the reconstructed images before any processing. The right column shows the reconstructed images after binary thresholding.

As is expected from the geometry of the system, the distance between the object and the sensor, D is the key parameter that affects imaging characteristics. If D is too small, then the field of view is expected to decrease because of loss of light (and hence, loss of information) from the off-axis points in the object. At the same time, if D is too large, then there is loss of light (and hence, loss of information) due to the sensor intercepting only a small fraction of the emitted light from each point on the object. As a result, one expects an optimal value of D for imaging. We determined this empirically by analyzing the performance of the sensor at D = 85mm, 165mm, 242mm, 343mm and 497mm. For each value of D, we obtained the matrix, *A*, calculated its singular values and also reconstructed a sample object (stickman) as summarized in Fig. 3(a). For each D, we also estimated the field of view by monitoring the extent of one PSF on the sensor plane. However, we used all the calibrated PSFs for reconstruction. We noticed that restricting the contributing points to within the field of view did not impact the final images. We were able to successfully recover objects with up to 32 LEDs in the horizontal direction and 30 LEDs in the vertical direction. The mounting platform blocked the bottom two rows of the LED matrix, which creates the discrepancy between the vertical and horizontal directions. Enhancement in LED discrimination was observed as a function of increasing distance and led to improved image reconstruction. For the largest value of D, it was observed that the quality of the reconstruction decreased slightly, possibly due to the reduction of SNR in the recorded image, as the amount of photon collected within the sensor surface decreases quadratically as a function of distance.

Next, we imaged lines oriented in the vertical, horizontal, and diagonal directions at the various sensor-matrix distances to access the effect of distance on the field of view and reconstruction quality. The results are summarized in Fig. 3(b). Reconstructions of vertical lines were better than those of the horizontal lines at all distances. All 30 LEDs (2 LEDs were blocked by our holder) were easily reconstructed in the vertical direction at all distances. On the other hand, at most only 26 LEDs were reconstructed properly in the horizontal direction. We attribute this asymmetry to the geometry of the sensor, which has 640 pixels in the vertical direction compared to 480 in the horizontal direction. Longer diagonal lines were reconstructed at larger values of D as indicated by the inset images in Fig. 3(b).

Lastly, we performed numerical analysis on calibration matrix *A* based on singular value decomposition. First, we compared singular-value-decay plot for the calibration matrix, *A* at each value of D. This plot is an indication of how ill-posed the inverse problem is[15]. Note that the singular-value-decay plot is computed from the same calibration dataset used to reconstruct the images. This plot shown in Fig. 3(c) shows that at D = 85mm, the singular values decay most quickly, which is consistent with our experimental results. Another way to estimate the difficulty of image reconstruction is by analyzing the correlation coefficient map (Fig 3d). Each element in the correlation matrix is the Pearson correlation coefficient between two calibration images, indicated by the vertical and horizontal indices. The horizontal matrix is created using a set of 32 LEDs along the horizontal line in the center. Vertical and diagonal matrices are created in the same manner. The correlation values of two sampled calibration images in any direction are always notably lower than 1, indicating that no two images within the calibration database closely resemble one another.

From this distance analysis, we determined that D=343mm is most appropriate for our setup. Reconstruction results presented in Figs. 2 and 4 were therefore acquired with calibration matrix collected at D=343mm. In addition to providing insight as to how the reconstruction performs at varying D, the aforementioned analysis implies the possibilities for computational refocusing and reconstruction of 3D information. Given the collection of the calibration database at multiple object distances, we could apply the same linear decomposition process used here to uncover the 3D spatial information of the scene. In this experiment, however, we restricted our experiments to the reconstruction on a single 2D plane, for simplicity.

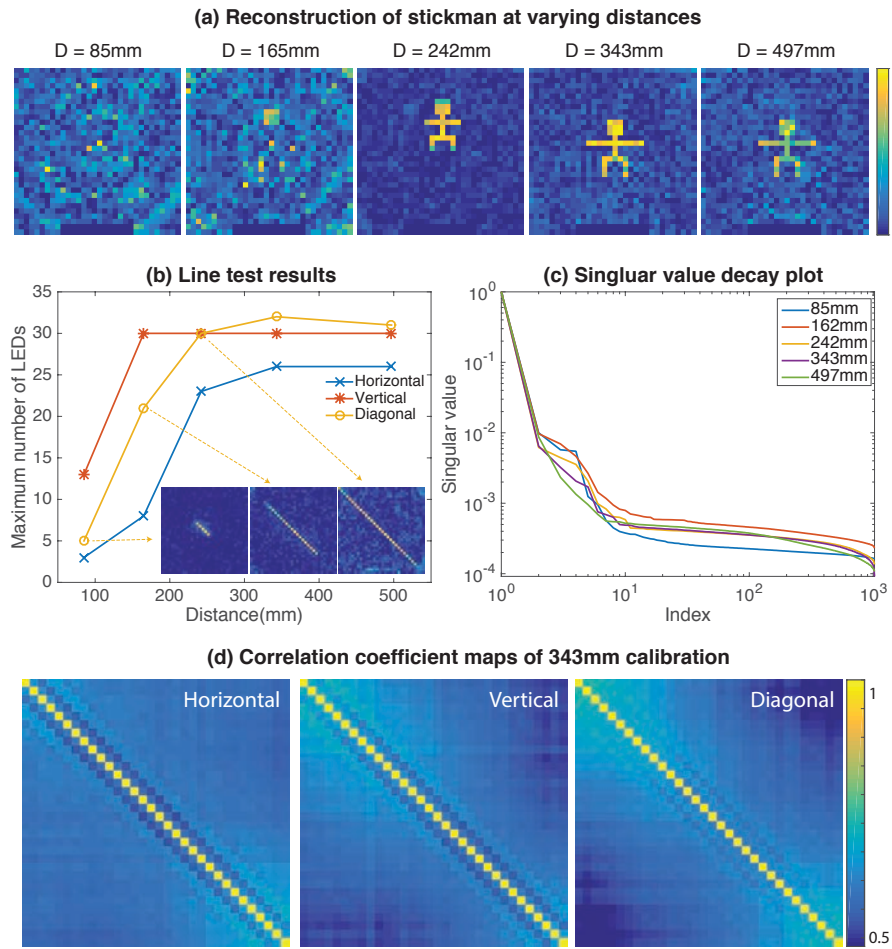

FIG. 3. Impact of object distance. (a) Reconstructed images obtained at various distances, D. (c) Length of a line object imaged at various distances, D in the horizontal, vertical and diagonal directions. Insets show reconstructed images of diagonal lines captured at the D = 85, 162, and 242mm. (d) Singular-value-decay plot of the calibration matrix, $A$ captured at various values of D. (e) Correlation matrix among the calibration images for D=343mm along the horizontal, vertical and diagonal directions.

Finally, we replaced the LED array with a liquid-crystal display (LCD) as indicated in Fig. 4(a). Using the calibration matrix obtained with the LED array, we were then able to reconstruct objects displayed on the LCD and these results are summarized in Fig. 4. In Fig. 4(a), 3 different geometries (all of color green, the same as the LEDs used for calibration) are shown. Fig. 4(b) shows a star pattern of various colors, all reconstructed computationally. It can be seen that the algorithm has difficulty reconstructing the red star. Note that all the reconstructions utilize the calibration data from the green LED array. In all cases, we note that reconstructions are possible, but of lower quality than in the case of the LED array. Finally, Fig. 4(c) shows how the distance D (between the LCD and the sensor) affect the imaging of an object comprised of 2 green letters, "HI." It is noteworthy that these reconstructions were all performed with the same matrix **A**, obtained at D = 343mm. The results in Fig. 4(c) indicate that the depth of field of the imaging system can be relatively large, although more experiments are required for precisely elucidating this effect. Nevertheless, these preliminary results indicate the feasibility of using just the image sensor for machine vision or similar applications, where anthropomorphic image quality is less important.

For different locations of LED to be discerned by the computational process, there must be distinct features in the raw image that uniquely differentiates each calibration image from another. In this experiment, we observed three locations on the sensor where small dust particles cast a scattering pattern onto the sensor. Therefore, the intensity pattern on the sensor shifts with the location of the LED. As an example, one element of the calibration matrix (the image of one LED at the top-left corner of the LED matrix) is shown in Fig. S1. In this frame, we noticed three regions with scattering patterns likely created by dust particles (indicated by the square outlines). If we closely observe one of these regions (denoted by the red square) for various frames (elements of A, corresponding to different LED positions in the matrix), we can see that the pattern shifts with the position of the LED in the matrix (bottom row). This would be then be consistent with the well-known pinspeck camera[16]. However, if we closely examine the same regions in the raw image of the extended object (with multiple LEDs), it can be seen that these scattering patterns disappear below the level of human perception as expected from the incoherent superposition of these patterns from multiple LEDs (see Figs. S2-S4 and supplementary information[17]). Therefore, our imaging system clearly differ from conventional pinspeck camera.

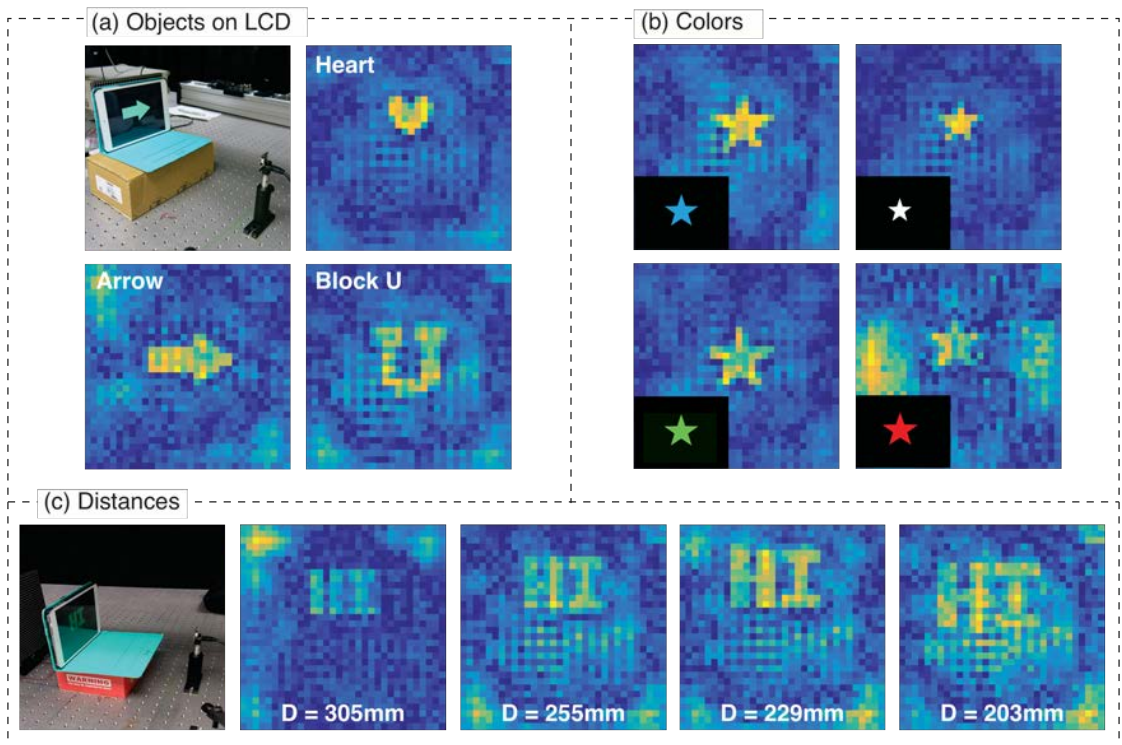

FIG. 4. Reconstruction of objects on LCD (iPad Mini, Apple) screen. (a) Photograph of setup and reconstruction results at D = 343mm. (b) Reconstruction of a star in four different colors. Insets show original displayed images. (c) Reconstruction of word 'HI' at varying object locations. In all cases including (c), the same matrix, **A** for D = 343mm were used.

Our hypothesis is that the image reconstruction is successful because of the combined effect of the subtle intensity variation across the entire sensor introduced by the diffraction and scattering caused by subtle defects and irregularities on the cover glass, as well as the few (in our case, 3) visible scattering patterns within the field of view of the sensor. It is important to note that clearly scatterers could be directly engineered onto the sensor to enhance such reconstructions in the future, which would make it similar to but much more compact than mask-based lensless imaging methods. Such techniques could obviously be extended to non-imaging computational problems including inference using deep learning and related algorithms[18-19].

In this paper, we described proof-of-concept experiments that demonstrate that lensless photography is possible with only an image sensor. This is achieved via the numerical solution of an inverse problem assuming incoherence of the object and linearity of the image-formation process. The technique is applied to both still-frame and video imaging. By calibrating the system at various planes, we show that it is possible to numerically refocus and achieve quasi-3D imaging as well. In other words, our only-sensor camera is always in focus (within limits of SNR). The technique is readily extended to color imaging. Further studies are necessary to understand the trade-offs related to resolution, field of view, depth of field and other metrics related to photography.

**Funding.** We gratefully acknowledge funding from the National Science Foundation (Grant # 10037833) and the Utah Science Technology and Research Initiative. K.I was supported by a University of Utah Nanotechnology Training Program fellowship. R.P was supported by University of Utah's Undergraduate Research Opportunities Program.

**Acknowledgments**. We thank Peng Wang for his help with the regularization algorithm, and Apratim Majumder for thoughtful discussions.

Supplementary Information for

# Lensless Photography with only an image sensor

Ganghun Kim,[1] Kyle Isaacson,[2] Rajesh Menon[1,a)]

[1]*Department of Electrical and Computer Engineering, University of Utah, Salt Lake City, 84111, USA*

[2]*Department of Bioengineering, University of Utah, Salt Lake City, 84111, USA*

*Corresponding author: rmenon@eng.utah.edu

### 1. Dust and particle shadow

Following four figures are enlarged copy of CMOS images presented in the main text. We marked areas where shadow appeared during the calibration procedure. Shadow completely disappears when an arbitrary object was drawn, while reconstruction was successful. Therefore, we concluded the shadow plays minor role in encoding spatial information. Images are shown in landscape orientation to maximally fit the image.

(a) Full calibration image

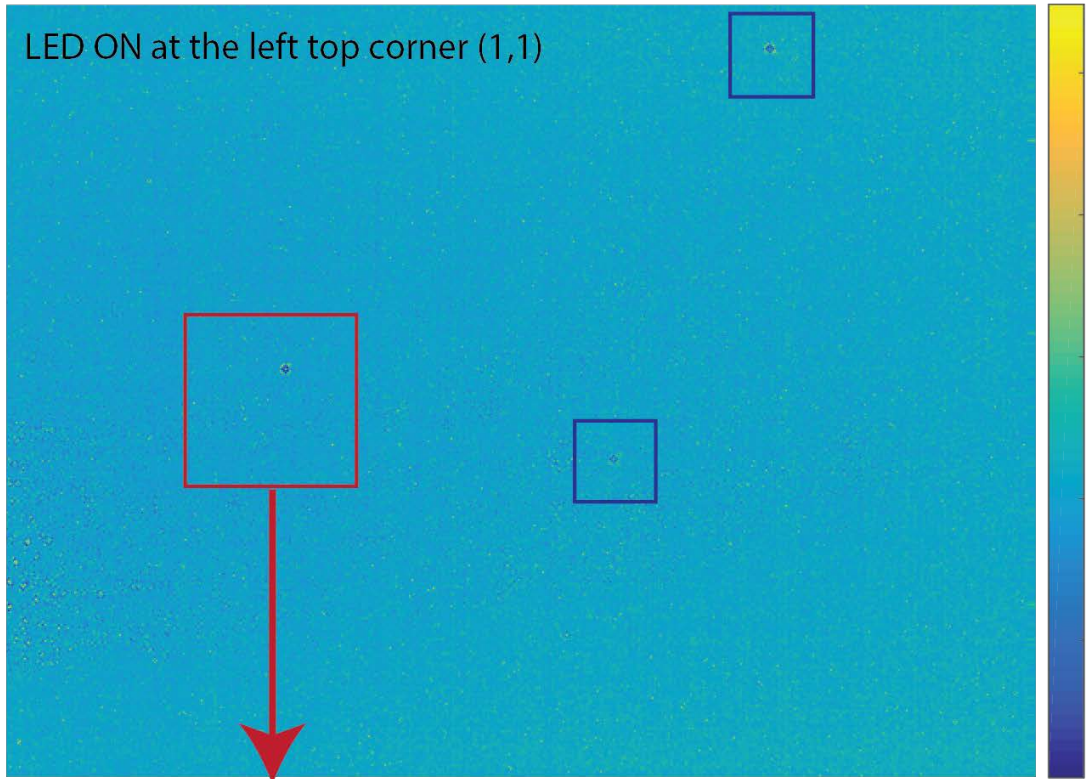

(b) Cropped image : shadow shifts with LED location

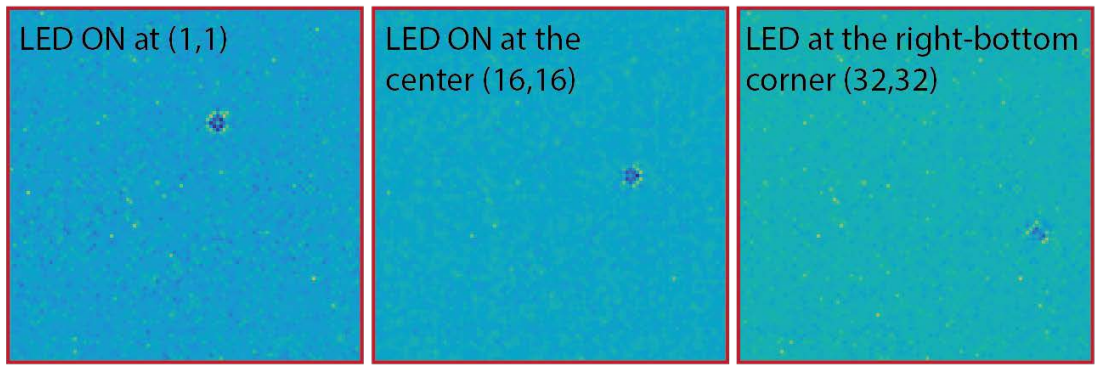

Figure S1. Detailed image of one LED located at the top-left corner shows 3 defects (possibly dust particles) on the sensor. These are indicated by the square outlines. The region marked with the red square is shown in the bottom for different LEDs, clearly indicating a shift of the "defect" scattering pattern.

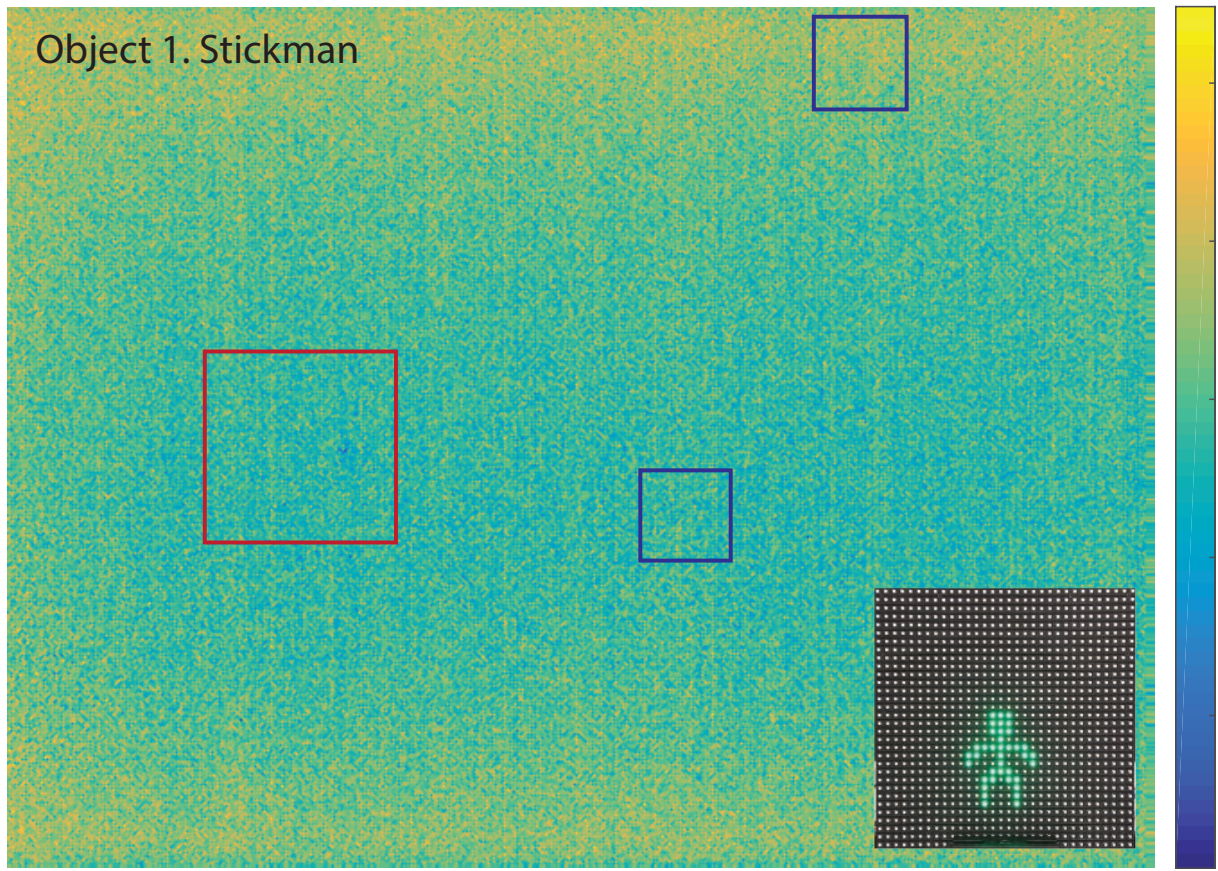

Figure S2. Enlarged CMOS image of the stickman object. Photograph of the LED is shown in the inset.

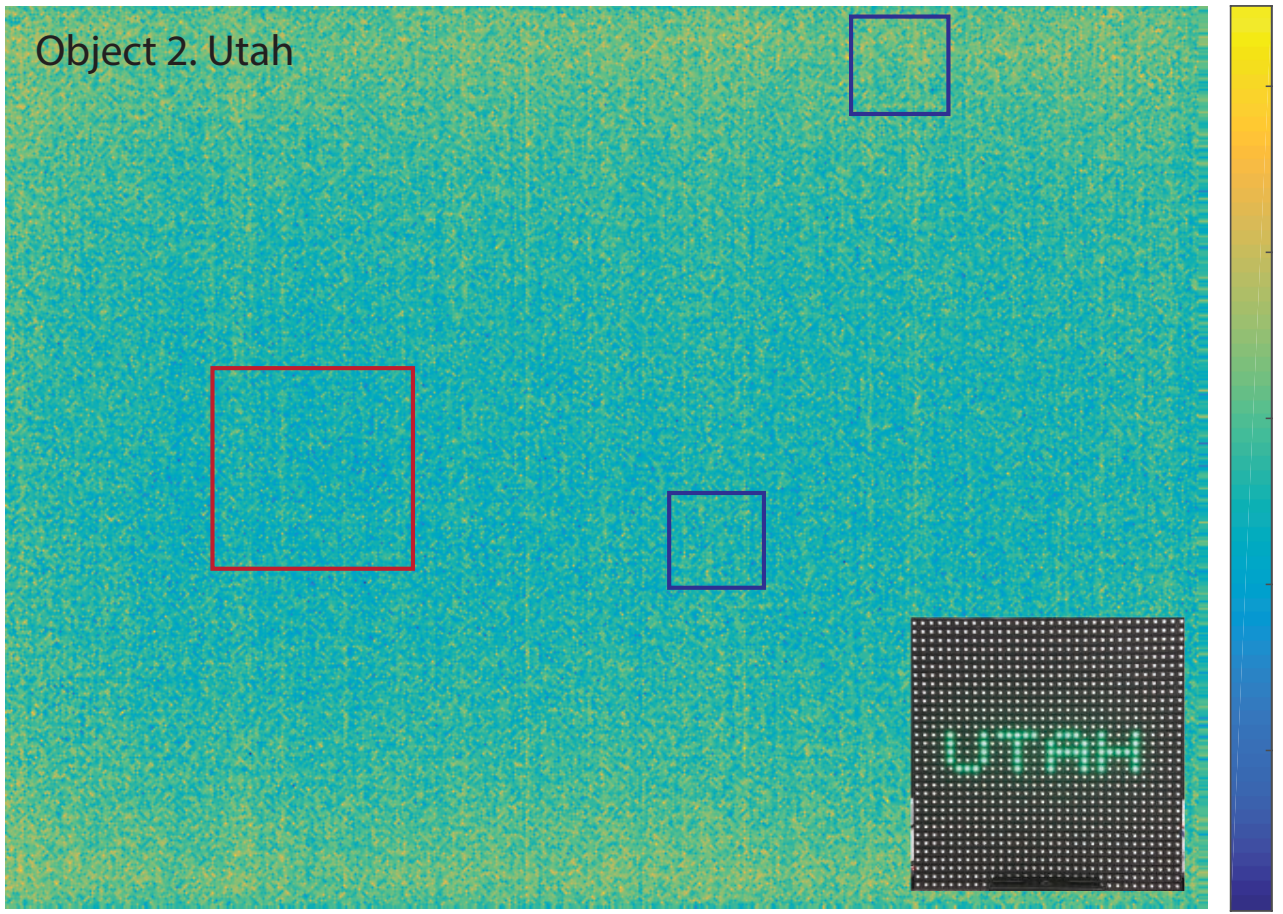

Figure S3. Enlarged CMOS image of the UTAH object. Photograph of the LED is shown in the inset.

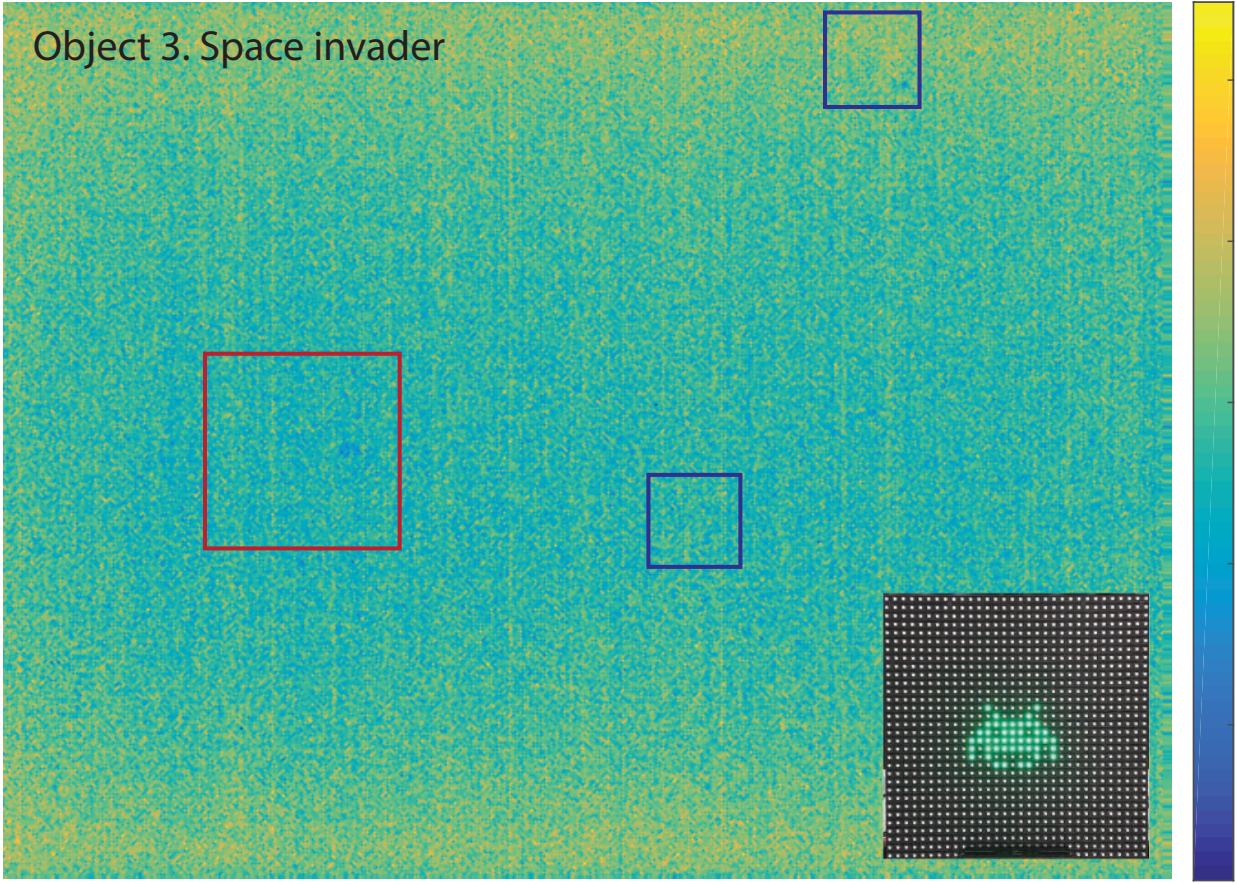

Figure S4. Enlarged CMOS image of the space invader object. Photograph of the LED is shown in the inset.

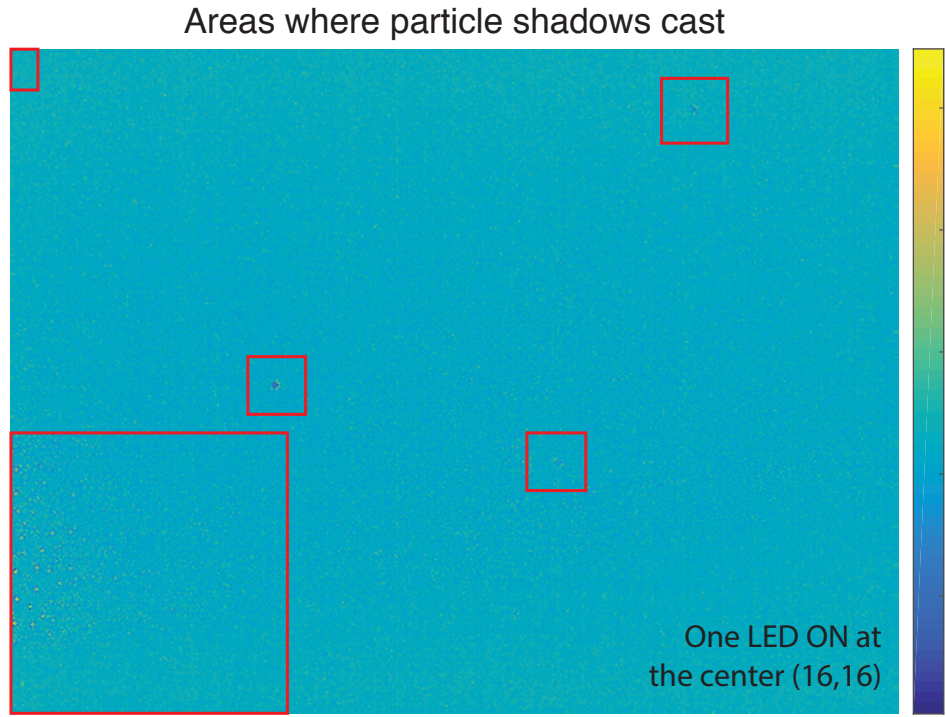
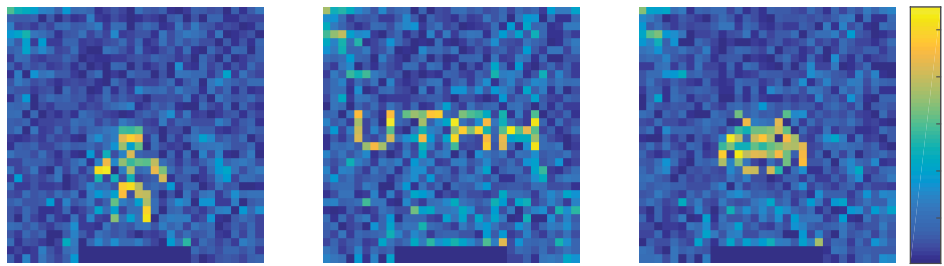
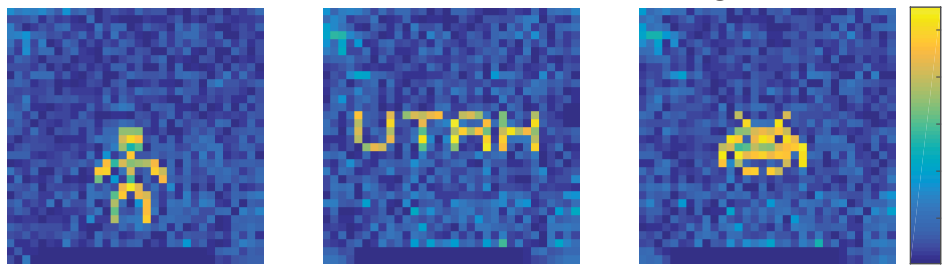

Figure S5. Reconstructions are performed after removing sections of images where particle shadows are casted both from the sensor data. Removed regions are marked by red boxes drawn on one of the calibration image. Results show successful reconstruction of the object without particle shadows, although with increased noise level.

2. **Imaging with CCD sensor**

Same imaging technique can be easily extended to other sensor type. We replaced our CMOS sensor with a CCD sensor (Clara Interline CCD, Andor) and successfully reconstructed objects displayed on LED panel.

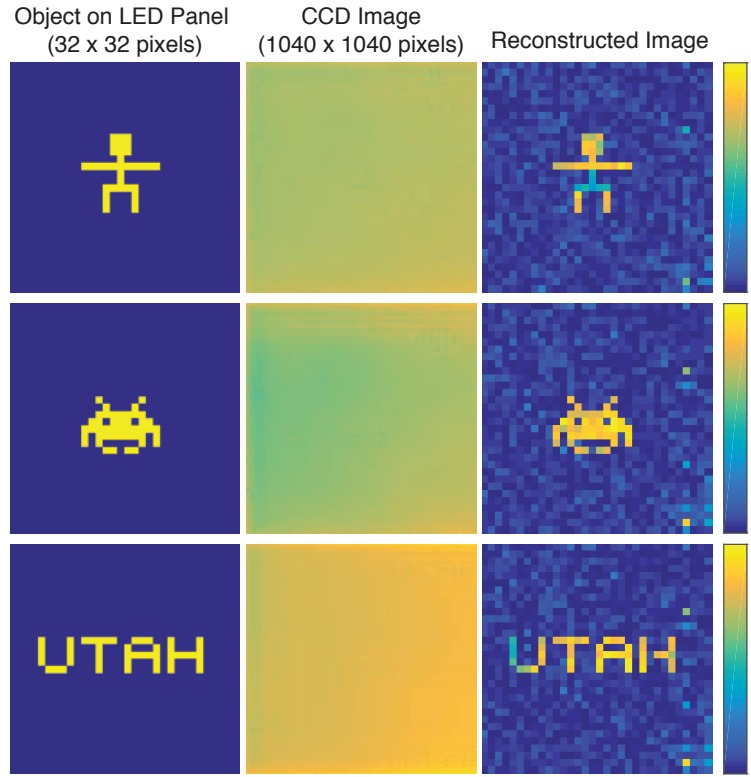

Figure S6. Lensless imaging performed using CCD sensor array. Objects similar to ones used in CMOS imaging were reconstructed. CCD sensor we used offers higher dynamic range and lower electrical noise, which we believe lead to better reconstruction quality.

3. **Supplementary Video 1 shows the animation of a jumping stick-man. Left: original scene. Center: Raw sensor video. Right: Reconstructed video.**